\documentclass[a4paper,12pt,onecolumn]{article}

\usepackage{FAS}
\usepackage[latin1]{inputenc}
\usepackage[misc]{ifsym}

\usepackage{cite}
\usepackage{amsmath,amssymb,amsfonts}
\usepackage{algorithmic}
\usepackage{graphicx}
\usepackage{caption}
\usepackage{textcomp}
\usepackage{xcolor}

\usepackage[export]{adjustbox}
\usepackage{url}
\usepackage{balance}
\usepackage{academicons}
\usepackage{booktabs}
\definecolor{orcidlogocol}{HTML}{A6CE39}
\definecolor{lime}{HTML}{A6CE39}
\DeclareRobustCommand{\orcidicon}{%
    \begin{tikzpicture}
    \draw[lime, fill=lime] (0,0) 
    circle [radius=0.16] 
    node[white] {{\fontfamily{qag}\selectfont \tiny ID}};
    \draw[white, fill=white] (-0.0625,0.095) 
    circle [radius=0.007];
    \end{tikzpicture}
    \hspace{-2mm}
}

\newcommand{\orcidWalter}{\href{https://orcid.org/0000-0003-4565-1272}{\orcidicon}}
\newcommand{\orcidRoss}{\href{https://orcid.org/0000-0001-8595-0379}{\orcidicon}}
\newcommand{\orcidXingcheng}{\href{https://orcid.org/0000-0003-1178-5221}{\orcidicon}}
\newcommand{\orcidRui}{\href{https://orcid.org/0000-0001-7359-1081}{\orcidicon}}
\newcommand{\orcidMarc}{\href{https://orcid.org/0009-0008-9223-2015}{\orcidicon}}
\newcommand{\orcidDaniel}{\href{https://orcid.org/0009-0008-5346-5473}{\orcidicon}}
\newcommand{\orcidAhmed}{\href{https://orcid.org/0000-0003-3702-8042}{\orcidicon}}
\newcommand{\orcidAkshay}{\href{https://orcid.org/0009-0000-4969-0119}{\orcidicon}}
\newcommand{\orcidTrivedi}{\href{https://orcid.org/0000-0002-0937-6771}{\orcidicon}}
\newcommand{\orcidKnoll
}{\href{https://orcid.org/0000-0003-4840-076X}{\orcidicon}}

\usepackage{tikz}

\usepackage{hyperref} 
\hypersetup{
    bookmarks=false,
    colorlinks=true,
    breaklinks=true,
    linkcolor=blue,
    filecolor=magenta,      
    hypertexnames=true,
    urlcolor=cyan,
    pagebackref=true,
    pdftitle={ICRA40},
    pdfpagemode=FullScreen,
    citecolor=green
    }
\begin{document}

\title{Enhancing Highway Safety: Accident Detection on the A9 Test Stretch
Using Roadside Sensors}
%

\author{
Walter Zimmer~$^{\text{\Letter}}$ \thanks{Chair of Artificial Intelligence and Robotics (AIR), Department of Computer Engineering (CE), School of Computation, Information and Technology (CIT), Technical University of Munich (TUM). $^{\text{\Letter}}$~Corresponding author: \texttt{walter.zimmer@cs.tum.edu}} \orcidWalter
Ross Greer~\thanks{Laboratory for Intelligent and Safe Automobiles (LISA) at the Uni. of California San Diego (UCSD).} \thanks{University of California Merced (UCM).}~\orcidRoss
Xingcheng Zhou~\footnotemark[1]~\orcidXingcheng
Rui Song~\footnotemark[1] \thanks{Fraunhofer Institute for Transportation and Infrastructure Systems (IVI).}~\orcidRui\\
Marc Pavel~\footnotemark[1] \orcidMarc \quad
Daniel Lehmberg~\footnotemark[1] \orcidDaniel \quad
Ahmed Ghita~\footnotemark[1]~~\thanks{SETLabs Research GmbH.}~~\orcidAhmed\\
Akshay Gopalkrishnan~\footnotemark[2] \orcidAkshay \quad
Mohan Trivedi~\footnotemark[2]~\orcidTrivedi \quad
\setcounter{footnote}{3}
Alois Knoll~\footnotemark[1] \orcidKnoll
}

%
%
\date{}

\makeatletter
\let\@oldmaketitle\@maketitle
\renewcommand{\@maketitle}{\@oldmaketitle
  \setcounter{figure}{0}    
  \vspace{-0.9cm}
  \centering
  \url{https://tum-traffic-dataset.github.io/tumtraf-a}\\[5pt]
  \includegraphics[width=1.0\textwidth,trim={0cm 6.6cm 0cm 0cm},clip,frame]{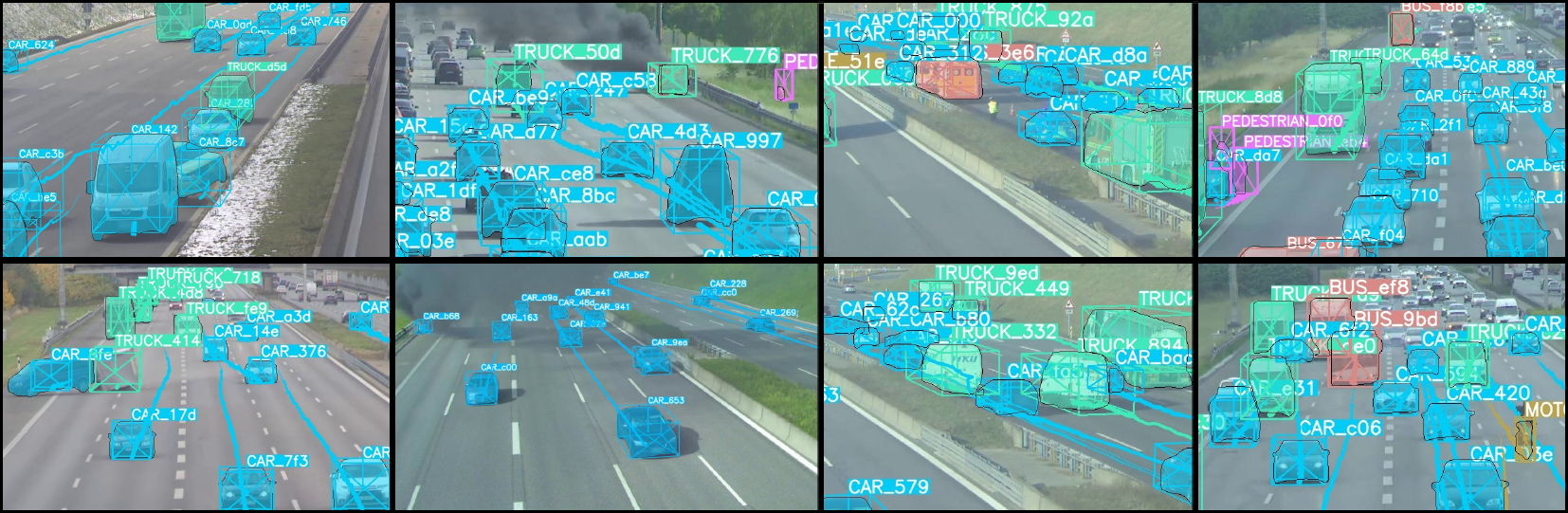}
  \vspace{-0.69cm}
  \captionof{figure}{
  Our dataset provides a visual representation of highway accidents with 3D box annotations, track IDs, instance masks, and vehicle trajectories. The accidents were captured using roadside cameras on the A9 Test Bed for Autonomous Driving. The scenes include crashes, overturned vehicles, and some instances of cars catching fire.
  }
  \label{fig:overview_figure}
  }
\makeatother

\maketitle

\thispagestyle{empty}
\vspace{0.7cm}
\begin{abstract}
Road traffic injuries are the leading cause of death for people aged 5-29, resulting in about 1.19 million deaths each year. To reduce these fatalities, it is essential to address human errors like speeding, drunk driving, and distractions. Additionally, faster accident detection and quicker medical response can help save lives. We propose an accident detection framework that combines a rule-based approach with a learning-based one.
We introduce a dataset of real-world highway accidents, featuring high-speed crash sequences. It includes 294,924 labeled 2D boxes, 93,012 labeled 3D boxes, and track IDs across 48,144 frames, captured at 10 Hz using four roadside cameras and LiDAR sensors. The dataset covers ten object classes and is released in the OpenLABEL format. Our experiments and analysis demonstrate the reliability of our method. 
\end{abstract}


\section{Introduction}

Data on rare events, such as long-tail events, is needed for improving machine learning models used in robotics, including perception, planning, and control \cite{christianos2023planning}. In autonomous driving, however, gathering this data is both expensive and challenging \cite{ghita2024activeanno3d, kulkarni2021create, fingscheidt2022deep}. Events like accidents and near-misses pose serious risks to human life and are difficult to capture naturally \cite{wang2022ips300+, greer2024and}.
Quick accident detection is critical because the time between a crash and medical response can determine survival. Automated accident detection can shorten this response time and help save lives.
Our dataset contributes to safer autonomous systems in several ways. First, it enhances learning for key perception tasks like detection \cite{liu2024graphrelate3d,carta2024roadsense3d,zimmer2023infradet3d,zimmer2023real,zimmer2022survey,zimmer2022realdomain,philipsen2015traffic,abualsaud2021laneaf,zimmer_pointcompress3d_2024,liu_graphrelate3d_2024,mohamed_transfer_2024}, tracking, segmentation \cite{greer2024patterns} or occupancy prediction \cite{song_collaborative_2024}. Second, its design enables cooperative perception, where roadside sensors capture the same scene from different angles to reduce blind spots. Finally, these sensors help create digital twins of traffic environments \cite{kremer2023digital}, extending visibility beyond a single vehicle's perspective. This broader view can improve early warnings and safer decision-making in risky traffic situations \cite{greer2023safe}.\\

\textbf{Our key contributions are the following:}
\begin{itemize}
\item We propose an accident detection framework capable of identifying and analyzing crashes and near-misses in real-time in various traffic scenarios.
\item We introduce a specialized highway accident dataset designed to capture rare and high-risk events, featuring 294,924 2D annotations and 93,012 3D annotations.
\item Our experiments and analysis demonstrate that our approach achieves top performance on this dataset.
\item We publicly release the dataset, detection framework, and development tools on our project website. 
\end{itemize}

\section{Related Work}
Accurate accident detection is crucial for improving road safety and emergency response times. However, existing methods have not been extensively tested on real traffic data from designated test sites. Real-world accident datasets \cite{xu2022tad} are rare and often lack sufficient data to train deep learning models effectively \cite{liu2024survey}. In this section, we categorize accident detection approaches into different groups based on their methodologies and data sources.

\subsection{Rule-Based Accident Detection}
Rule-based methods \cite{pirdavani_application_2015} rely on predefined traffic rules and heuristics to identify accidents. These approaches typically use sensor readings, such as abrupt changes in speed or sudden braking, to infer collisions. While effective in controlled environments, rule-based systems struggle with unseen accident scenarios and complex traffic settings.

\subsection{Learning-Based Accident Detection}
Machine learning models, particularly deep learning approaches, have been increasingly used for accident detection. These methods use large datasets to learn patterns associated with crashes \cite{zahid2024datadriven, adewopo2024smart}. Supervised learning techniques train models on labeled accident and non-accident data, whereas unsupervised approaches \cite{yao2023dota} detect anomalies that may indicate collisions. Despite their effectiveness, these models require extensive labeled data, which is often difficult to obtain.

\subsection{Trajectory-Based Accident Detection}
Trajectory-based methods \cite{yang_freeway_2021} analyze the motion patterns of vehicles to detect unusual behavior that may lead to accidents. These techniques use sensor data to track vehicles in 2D or 3D and identify sudden deviations, abrupt stops, or erratic lane changes. While useful for predicting potential accidents, trajectory-based methods may struggle with false positives and ID switches in highly occluded traffic scenarios.

\subsection{Vision-Based Accident Detection}
Vision-based approaches \cite{fang_vision-based_2024} utilize camera footage to detect accidents by identifying visual cues such as vehicle deformations, smoke, or fire. These methods often employ convolutional neural networks (CNNs) to classify accident and non-accident scenarios. Although vision-based detection can provide valuable real-time insights, it is sensitive to lighting conditions, occlusions, and camera angles.

\subsection{Video-Based Accident Detection}
Video-based methods \cite{maaloul_adaptive_2017} extend image-based approaches by analyzing temporal information in video streams. These techniques apply deep learning models such as recurrent neural networks (RNNs) or transformers to track accident progression over time. Video-based detection enhances accuracy by considering motion dynamics but requires substantial computational resources and well-annotated datasets.

\subsection{Accident Detection Using Synthetic Data}
To overcome the scarcity of real-world accident data, synthetic datasets have been developed for training and evaluating accident detection models. One notable example is DeepAccident \cite{wang2024deepaccident}, which contains 691 synthetic accident scenarios generated in the CARLA simulator \cite{dosovitskiy_carla_2017}. These accidents are based on crash reports from the National Highway Traffic Safety Administration (NHTSA) and include labeled data from four vehicles and one roadside infrastructure camera. However, a key limitation of DeepAccident is its reliance on simulated environments, which may not fully capture the complexities of real-world crashes. Addressing the sim-to-real gap is essential to improve the generalization of perception models trained on synthetic data.

\subsection{Accident Detection Using Real Data}
Real-world accident detection datasets are limited due to the difficulty of capturing and annotating crash events. Some existing datasets contain dashcam footage, roadside surveillance videos, or event logs from connected vehicles. However, many of these datasets lack sufficient labeled data to train deep learning models effectively. Furthermore, variations in road conditions, weather, and camera perspectives introduce additional challenges in real-world accident detection.

In summary, while significant progress has been made in accident detection, the scarcity of real-world datasets remains a major challenge. 
Related approaches may be used to detect both accidents and general traffic anomalies \cite{greer2024towards, greer2023pedestrian}. 
Existing methods often struggle with generalization due to limited real-world training data or the reliance on synthetic environments. To address this gap, our accident detection framework integrates both rule-based and learning-based approaches to reliably identify accidents in 3D. By leveraging real-world roadside sensor data and combining explicit traffic rules with deep learning models, our framework enhances detection accuracy, improves robustness across diverse traffic conditions, and enables faster emergency response times.

\section{Accident Detection Framework}
We propose an automatic accident detection framework that operates in real-time using roadside cameras. Our method combines both rule-based and learning-based approaches to improve accuracy and reliability. The framework first applies a rule-based approach to detect potential accidents based on vehicle trajectories. If an accident is identified, a learning-based model further verifies the event to minimize false positives.

\subsection{Rule-Based Accident Detection}
The rule-based approach relies on analyzing vehicle trajectories to detect unusual behavior indicative of an accident. Predefined thresholds, such as sudden stops, erratic lane changes, or prolonged stationary positions on active lanes, are used to classify accidents in real-time. This method provides immediate accident classifications for each vehicle in the current frame, enabling rapid initial detection.
The rule-based approach analyzes lane positions, distance measurements, and object speeds to detect potential accidents based on predefined maneuver rules:
\begin{align}
&velocity_i \geq \frac{15~km/h}{3.6}\label{eq:rba1}\\
&velocity_i > velocity\_lead_i \label{eq:rba2}\\
&velocity_i \geq velocity_j \quad \quad \forall \quad i < j \leq N\label{eq:rba3}\\
&distance\_lead_i \geq distance\_threshold\label{eq:rba4}\\
&distance\_lead_i < (\frac{( velocity_i - velocity\_lead_i )}{30}) ^ 2\label{eq:rba5}\\
&ttc\_leading_i \leq ttc\_threshold\label{eq:rba6}
\end{align}
If all six rules are met at the same time, the traffic participant is classified as being involved in an accident.

\subsection{Learning-Based Accident Detection}
Once a potential accident is flagged by the rule-based method, a learning-based approach refines the prediction. We train a YOLOv8 model \cite{yolov8_ultralytics} on our custom accident dataset to detect accidents directly from camera images. Here we use the CVAT annotation tool \cite{cvatai_corporation_computer_2023} and annotate accident events with 2D bounding boxes. We split the full dataset into a training (80\%), validation (10\%), and test (10\%) set. It contains 3,725 image frames in total: 2,896 for training, 410 for validation, and 419 for testing. 
An accident event is visible in 59.48\% of images.
Table \ref{tab:customDS} provides an overview of the dataset composition. \\

\begin{table}[htb]%
    \centering%
    \begin{tabular}{l r r r r}
        \toprule
        \textbf{Dataset} & \textbf{Train} & \textbf{Validation} & \textbf{Test} & \textbf{Total} \\
        \midrule
        Accident frames & 1,799 & 224 & 292 & 2,216 \\
        Non-accident frames & 1,097 & 186 & 127 & 1,509 \\
        \midrule
        Total & 2,896  & 410  & 419  & 3,725  \\
        \bottomrule
    \end{tabular}
    \caption{Total number of image frames in the custom dataset. 59.48\% of images capture an accident event.}
    \label{tab:customDS}%
\end{table}

The model filters detections based on a confidence threshold of 0.8. To further reduce false positives, an accident must be detected in at least three consecutive frames before confirmation. Additionally, we aggregate accident detections from multiple roadside cameras to enhance detection accuracy and reduce occlusion effects.

\section{Evaluation}
   
\subsection{Dataset}
Our dataset features 48,144 labeled camera and LiDAR frames with 294,924 2D and 93,012 3D box annotations, tracking labels and trajectory information, and classification of ten different instance types, including cars, trucks, buses, trailers, vans, pedestrians, motorcycles, bicycles, emergency vehicles, and others. The data is captured from roadside cameras and LiDARs during day and nighttime and labeled with the \textit{3D Bounding Box Annotation Toolbox} (3D BAT) \cite{zimmer20193d}. Featured accidents include instances of high-speed lane changes with failure to notice stopped traffic, overturning of vehicles during collision, vehicles catching fire, a variety of emergency response vehicles, and more. Some example cases are illustrated in Fig. \ref{fig:overview_figure}.

Our dataset can be used as ground truth for the development and verification of AI-based detectors, tracking, fusion algorithms, trajectory prediction, and to understand and analyze the occurrence and the after-effects of naturally occurring high-speed crash incidents and other accidents on the autobahn.

The dataset, as well as its preceding datasets \cite{cres_a9-dataset_2022,zimmer_tumtraf_2023,cress2024tumtrafevent,zimmer2024tumtraf,zhou_warm-3d_2024}, are available for academia and industry and include a development kit repository for ease of use \cite{zimmer2024devkit}. 

\subsection{Experiments}
To evaluate our framework, we recorded camera images and processed fused perception results over 128 days, storing the data in rosbag files. The automatic accident analysis was performed on 12,290 15-minute video segments, identifying 831,969 unique vehicles. Our framework detected: 3,748 standing vehicles in active driving lanes, 138 standing vehicles in shoulder lanes, 120 breakdown events, and one accident.

Moreover, we evaluate the accuracy and runtime performance of both accident detection models on our dataset. 
The rule-based approach needs 10.41 ms per frame (95.05 FPS) on an NVIDIA RTX 3090 GPU. The total processing runtime for a 15-minute rosbag file with 22,500 ROS messages recorded at 25 FPS is 234.25 seconds. This includes the calculation of the lane ID, the distance calculation between all vehicles, and the scenario classification.

Our experiments (see Table \ref{tab:experiments}) show that the learning-based accident detection method achieved a precision of 0.8 and a perfect recall of 1.0 on the test data, ensuring high accuracy in identifying accidents. Additionally, the detection process takes 10.41 ms for the rule-based approach and 16.13 ms (using TensorRT acceleration) for the learning-based approach, making it efficient for real-time applications.

\begin{table}[h]
    \centering
    \begin{tabular}{lrrrrr}
        \toprule
        \textbf{Approach} & \multicolumn{4}{c}{\textbf{Accuracy}} & \textbf{Runtime [ms]}\\
        \cmidrule{2-4}
        & \textbf{Precision} & \textbf{Recall} & \textbf{F1-Score} & \textbf{AP}\\
        \midrule
        Rule-based Approach &  \textbf{1.000} & 0.500 & 0.667 & \textbf{0.97} & 10.41 \\
        Learning-based Approach & 0.800 & \textbf{1.000} & 0.889 & 0.75 & \textbf{16.13}\\
        \bottomrule
    \end{tabular}
    \caption{Evaluation of the accident detection framework on the test set.}
    \label{tab:experiments}
\end{table}

\begin{figure}[h]
    \centering
    \includegraphics[width=0.32\linewidth,frame]{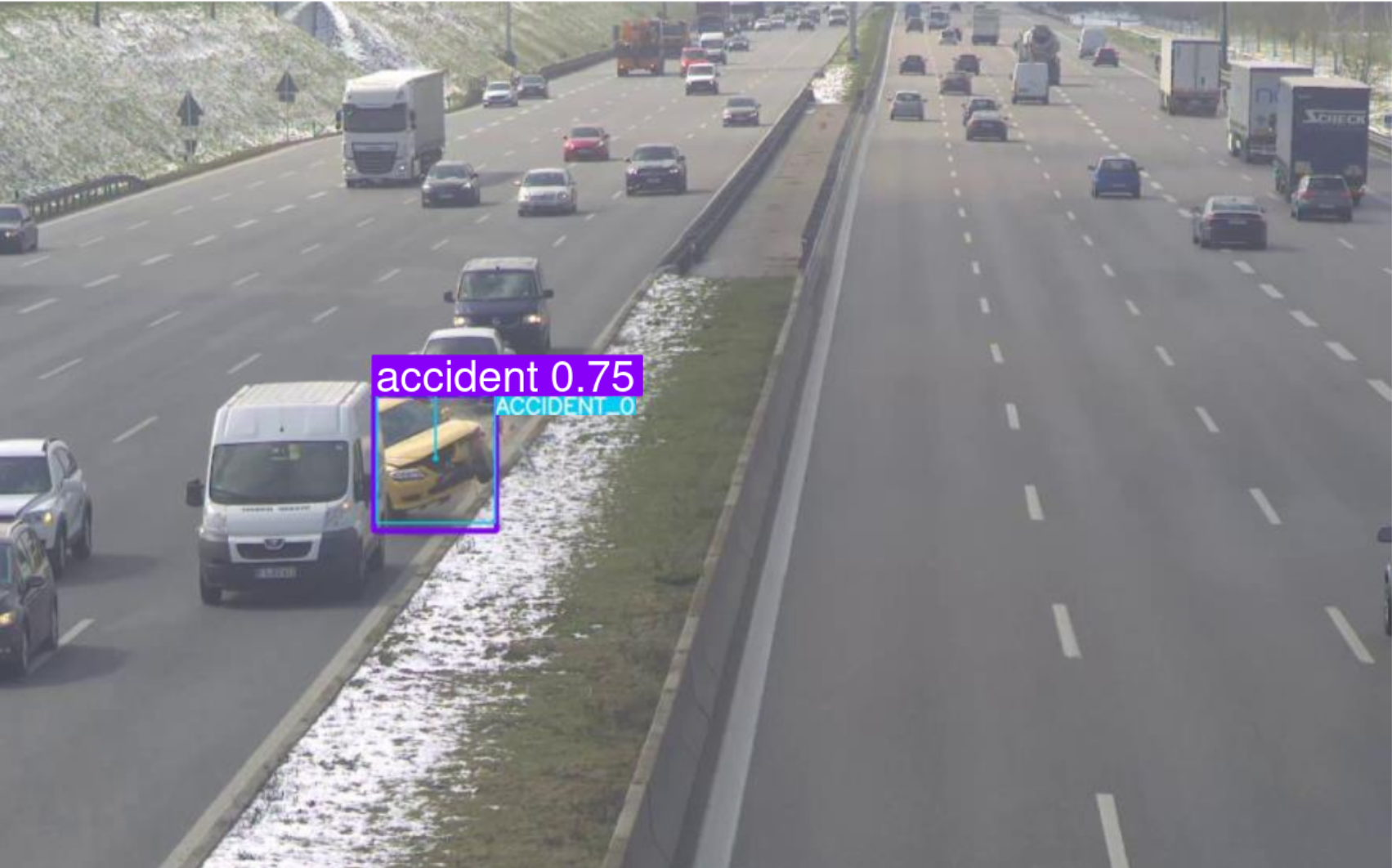}
    \includegraphics[width=0.32\linewidth,frame]{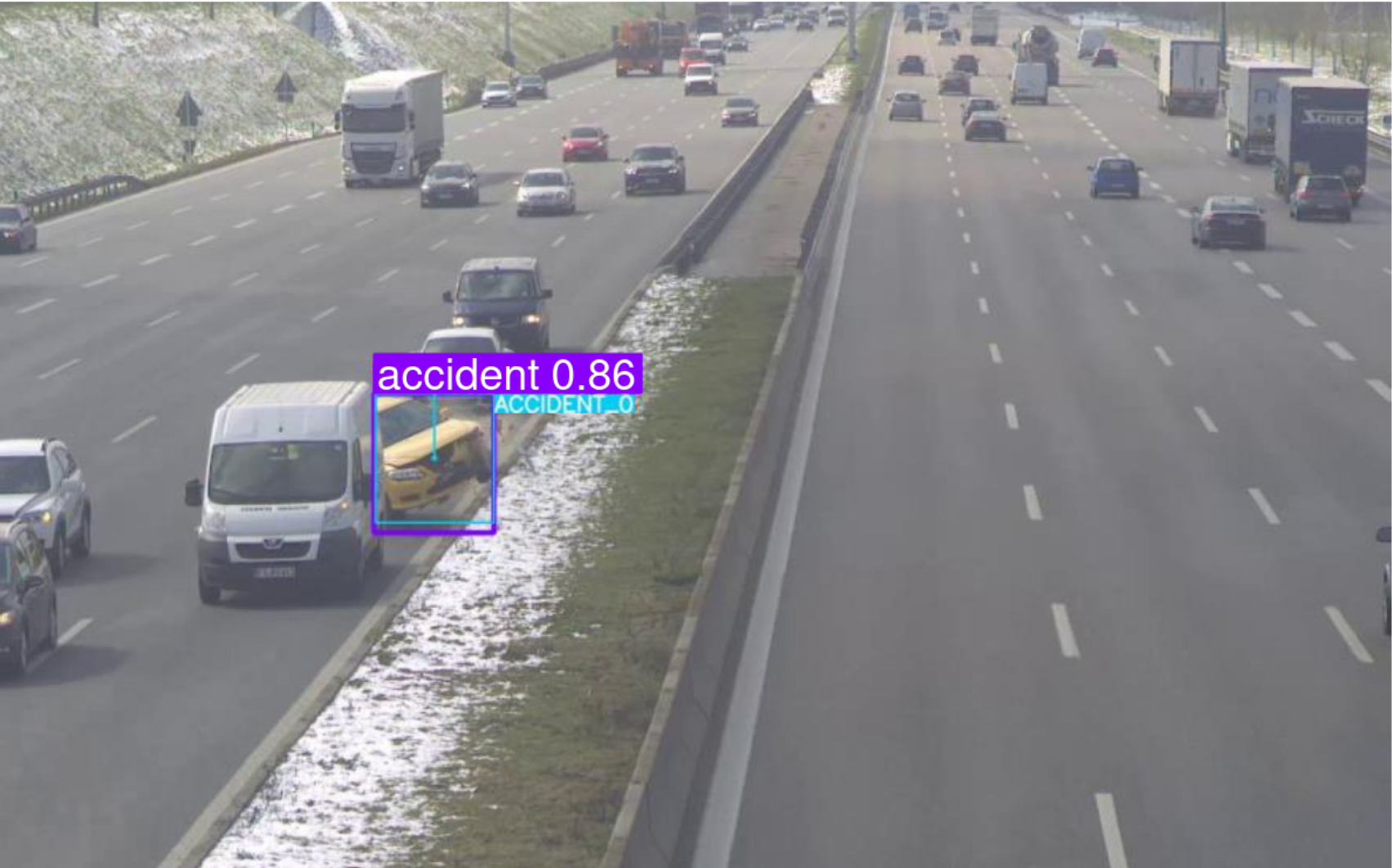}
    \includegraphics[width=0.32\linewidth,frame]{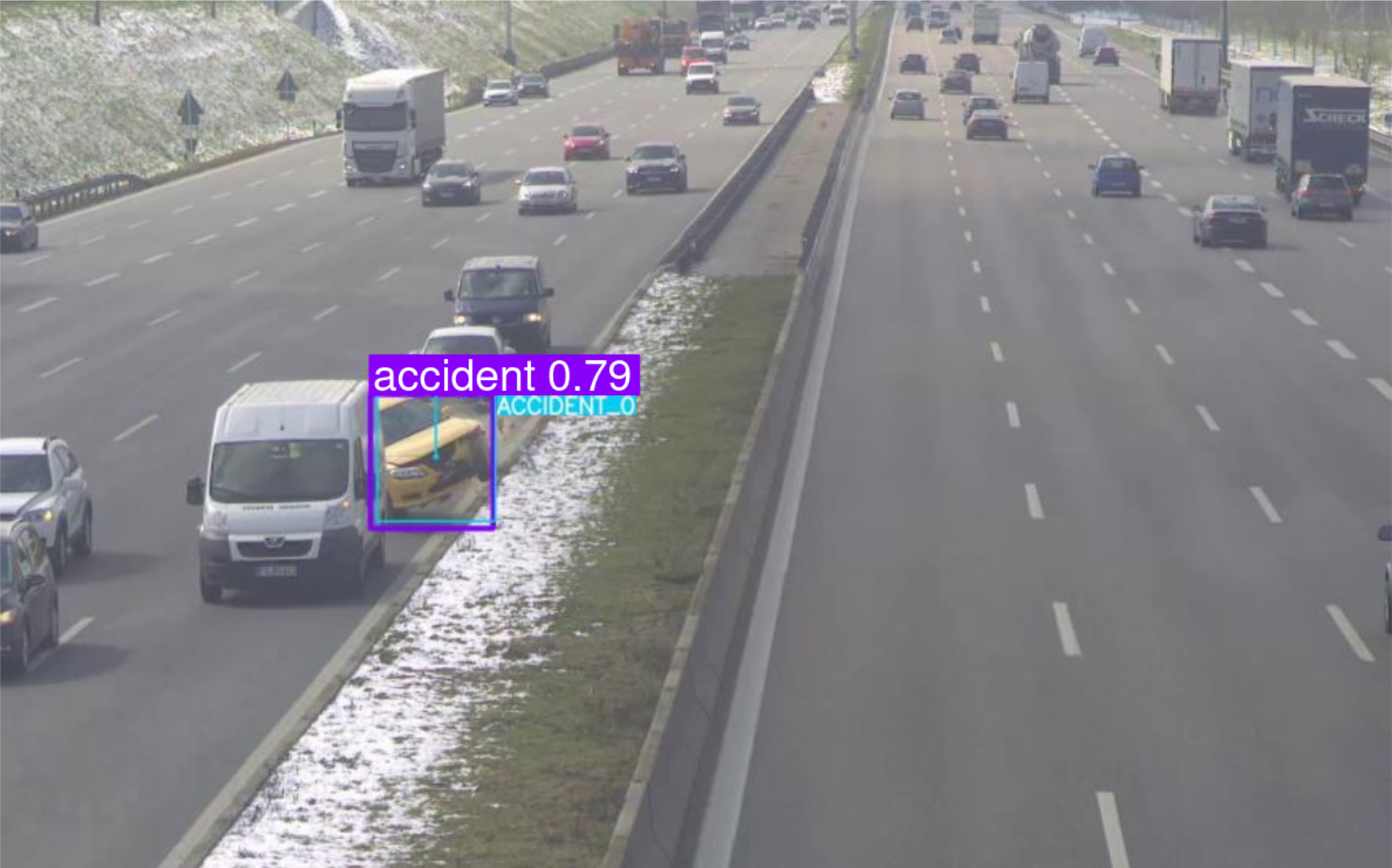}
    \caption{Visual comparison of the accident detection results (purple) on the test set of our accident dataset. Labeled accidents are displayed in turquoise. This test sequence shows a car accident scenario with a yellow car and a white van. We train object detection models with different input image resolutions (1920 px, 1280 px, and 960 px) and find out that an image resolution of 1280 px works best.}
    \label{fig:qualitative_results}
\end{figure}

Qualitative results of our accident detection framework are shown in Fig. \ref{fig:qualitative_results}. 
We train object detection models using image resolutions of 1920 px, 1280 px, and 960 px, determining that 1280 px yields the best performance.

\section{Conclusion}
Traffic accidents remain a leading cause of fatalities, emphasizing the need for rapid detection to enhance emergency response times. In this work, we introduce a real-world highway accident dataset and propose a hybrid accident detection framework that integrates rule-based and learning-based approaches. Our method demonstrates strong performance in real-time accident detection using roadside infrastructure sensors, effectively identifying crashes and near-miss events. Experimental results confirm the reliability and efficiency of our approach, highlighting its potential to improve traffic safety and emergency response. Future research should focus on enhancing model robustness, expanding predictive capabilities, and bridging the gap between synthetic and real-world accident data.

\section{Future Work}
While our framework demonstrates strong performance in real-time accident detection, several areas remain for future improvement. One key challenge is bridging the gap between synthetic and real-world data. Future research should explore domain adaptation techniques and self-supervised learning to enhance model generalization across different environments. Another important direction is improving model robustness under varying traffic conditions, such as heavy congestion, adverse weather, and nighttime scenarios. This could involve integrating multi-modal sensor data, such as LiDAR and radar, to complement camera-based detection. \\
We also plan to extend our rule-based approach to detect lateral collisions and more complex accident scenarios, such as multi-vehicle crashes and near-miss events. Additionally, refining our learning-based model with larger and more diverse datasets could further reduce false positives and improve real-time decision-making.\\
In the future, we plan to use Vision-Language Models (VLMs) \cite{zhou_vision_2024} to enhance the accuracy and reliability of accident detection. Finally, expanding our framework to incorporate predictive accident detection, identifying high-risk situations before a collision occurs, could significantly enhance traffic safety. This may also involve trajectory forecasting.\\
By addressing these challenges, we aim to develop a more reliable and comprehensive accident detection system that can be deployed in real-world autonomous driving and intelligent transportation applications.

\section{Acknowledgments}
This research was supported by the Federal Ministry of Education and Research in Germany as part of the AUTOtech.agil project \cite{van_kempen_autotechagil_2023} (Grant Number: 01IS22088U). We thank our collaborators and institutions for their valuable contributions and support. Additionally, we appreciate the efforts of the research community in advancing accident detection and traffic safety technologies.
\balance
\bibliographystyle{ieeetr}
\bibliography{main}
\end{document}